**REVIEW ARTICLE**    OPEN

# ChatGPT for shaping the future of dentistry: the potential of multi-modal large language model

Hanyao Huang [1]✉, Ou Zheng [2]✉, Dongdong Wang[2], Jiayi Yin[1], Zijin Wang[2], Shengxuan Ding[3], Heng Yin[1], Chuan Xu[4,5], Renjie Yang[6], Qian Zheng[1] and Bing Shi[1]

The ChatGPT, a lite and conversational variant of Generative Pretrained Transformer 4 (GPT-4) developed by OpenAI, is one of the milestone Large Language Models (LLMs) with billions of parameters. LLMs have stirred up much interest among researchers and practitioners in their impressive skills in natural language processing tasks, which profoundly impact various fields. This paper mainly discusses the future applications of LLMs in dentistry. We introduce two primary LLM deployment methods in dentistry, including automated dental diagnosis and cross-modal dental diagnosis, and examine their potential applications. Especially, equipped with a cross-modal encoder, a single LLM can manage multi-source data and conduct advanced natural language reasoning to perform complex clinical operations. We also present cases to demonstrate the potential of a fully automatic Multi-Modal LLM AI system for dentistry clinical application. While LLMs offer significant potential benefits, the challenges, such as data privacy, data quality, and model bias, need further study. Overall, LLMs have the potential to revolutionize dental diagnosis and treatment, which indicates a promising avenue for clinical application and research in dentistry.

## INTRODUCTION

Artificial intelligence (AI) has promoted recent progress in digital health for many years.[1,2] AI-equipped applications in dentistry have been found useful in analyzing medical imaging, including diagnosing dental caries,[3,4] periodontitis,[5] and implantitis,[6] and assisting oral and maxillofacial surgery with surgical planning.[7] Besides the imaging data, audio data analyses can also benefit from deep-learning applications, as speech is one of the most important functions of the oral structure.[8,9] Furthermore, dental education is another emerging application.[10] GPT-4, released by OpenAI, embarks on a new period of AI-powered large language models (LLMs). ChatGPT, built upon GPT-4, stirred up lots of interest among millions of scientists and engineers on account of its impressive human conversational response as a chatbot.[11] However, its potential impact on revolutionizing a series of technologies is more significant. Unlike earlier applications, ChatGPT is cultivated conversationally upon a tremendous knowledge base, enabling informative communications for the improvement of decision knowledge. Before ChatGPT, most AI technologies focus on the system of one input and one output, which relies on the amount of training data. With the influx of new data, re-training is required to update the existing model for more accurate decision-making. ChatGPT breakthroughs this mode and incorporates conversation to dynamically capture multiple sources of existing knowledge for question answering.[2,12,13] This human-friendly feature facilitates the diagnosis process and causes a significant change in the status quo, and its advancement will also shape digital health in dentistry.[1,14–16] The purpose of this paper is to provide an overview of the potential application of ChatGPT in dentistry.

## JOURNEY OF LLMS

Before LLMs garner significant attention, language modeling has undergone a series of revolutions in the past decade. The early natural language model is carried out with n-gram modeling,[17] which is probabilistic modeling yet effective for medical research.[18,19] The first milestone work after n-gram modeling is word embedding, which represents words in vector space to understand the natural language from a new quantitative perspective, promoting clinical research on document analysis.[20,21] Among a range of representation modeling, ELMo[22] proposed by AllenNLP changes the game to a bi-direction model pretraining. This modeling approach also influences medical language research[20] and is also evaluated.[23] Since then, bi-directional deep-learning models have been proposed like BERT[24] and Generative Pretrained Transformer (GPT).[25] Built upon these models, a range of medical language models are proposed to accelerate medical research progress, such as a family of medical BERT models,[26–29] and clinical researchers found that the increase in model size significantly improves a variety of medical applications.[21,30–32] However, they are limited to medium model

¹State Key Laboratory of Oral Diseases & National Clinical Research Center for Oral Diseases & Department of Oral and Maxillofacial Surgery, West China Hospital of Stomatology, Sichuan University, Chengdu, China; ²Department of Civil, Environmental & Construction Engineering, University of Central Florida, Orlando, USA; ³College of Transportation Engineering, University of Central Florida, Orlando, USA; ⁴School of Transportation and Logistics, Southwest Jiaotong University, Chengdu, China; ⁵C2SMART Center, Tandon School of Engineering, New York University, Brooklyn, USA and ⁶State Key Laboratory of Oral Diseases & National Clinical Research Center for Oral Diseases & Eastern Clinic, West China Hospital of Stomatology, Sichuan University, Chengdu, China
Correspondence: Hanyao Huang (huanghanyao_cn@scu.edu.cn) or Ou Zheng (ouzheng1993@knights.ucf.edu)
These authors contributed equally: Hanyao Huang, Ou Zheng

Received: 26 March 2023 Revised: 6 July 2023 Accepted: 13 July 2023
Published online: 28 July 2023

scales due to architecture design and hardware support, although some efficient algorithms are proposed for the medical domain.[20,33,34] One of the most important LLMs is T5, with 11 billion parameters proposed by Google.[35] Another rival model is GPT-3, developed by OpenAI, which contains 175 billion parameters. These billion-parameter models embark on a new chapter of LLMs and their applications. One of the most successful application instances is ChatGPT, a variant of InstructGPT[36] developed upon GPT-3, optimized by conversational response training. ChatGPT is equipped with interactive training which involves human feedback reinforcement learning and exhibits powerful language skills to generate human-like texts in real-time conversation. This interactive modeling also influences medical research like education.[37,38] All these rely on large-scale representation pretraining, which becomes critical to complex problem-solving with data in cross-modality, even for ChatGPT.

Large-scale vision-language pretraining
Vision-language pretraining is an important approach to solving text-to-image or image-to-text tasks, which trains a deep neural network with large image and text datasets. One of the vital training frameworks is CLIP, proposed by OpenAI, which is further improved by Salesforce to BLIP.[39] For text-to-image, GAN as an image generation prototype model can be integrated with text representation to generate diverse, authentic-looking but synthetic images.[40] Recently, the Diffusion Model,[41] a rival model of GAN emerging with higher computation efficiency and image diversity, has been incorporated for vision-language representation pretraining. For example, DALL-E v2[42] leverages CLIP ranked representation and diffusion model to generate image understanding sentences. To address medical domain-specific problems, a series of efficient representation learning models are also developed to empower intelligent healthcare services. BERT provides an effective solution to efficient inference and analysis of disease.[43] Multi-modal learning is also considered to improve medical visual-questioning-answer processes.[44] Although various applications are proposed, the study on how to integrate this powerful model with dental diagnosis is still limited.

Large-scale audio-language pretraining
Compared to vision-language models, audio-language pretraining does not prevail, but the representation learning with audio and text data still exhibits impressive audio-to-text performance. Some medium-scale models like MusCALL,[45] CTAL,[46] Wav2Seq,[47] and LAVA[48] indicate the superiority of representation pretraining on speech recognition. One of the important large-scale representation models is Whisper,[49] released by OpenAI, which is trained on 680,000 h of diverse audio-text pairs from the web. Inspired by the success of these works, an improved medical speech-to-text pretraining model is developed to more effectively link vocal signals to language generation and understanding.[50] Since audio-language pretraining research is still under exploration, the limited study demonstrates how to employ this pretraining framework to facilitate oral treatment.

Multi-modal LLM
With tremendous success in cross-modal training, more research attempts to incorporate multi-modal representation learning to empower LLMs. As one of the successful attempts, GPT-4[51] demonstrates the competence of LLMs in a multitude of NLP applications, such as higher scores in GRE tests and other question-answering tasks. This implies a higher potential for Multi-Modal LLM in various areas, such as digital health. For example, multi-modal learning is conducted to facilitate medical services, which incorporates images, audio, and texts into training for a more comprehensive and robust model.[43,45] However, due to limited data availability, more research still attempts to explore the merit of multi-model LLM for medical fields, especially dental clinic research.

LLM as a ubiquitous solution
As LLMs become increasingly widely recognized, more representations will be embedded into the models to enhance their general problem-solving skills. The training process with a larger scale of data will yield a ubiquitous solution to problems of all kinds. For example, ChatGPT has served as a valuable tool to assist medical education for more effective instruction and analysis of teacher-student interaction.[37,38] Medical writing can be assisted or even accomplished by ChatGPT,[32] which enables efficient documentation. Language challenges in medical research or clinical processes can also be alleviated by ChatGPT.[52]

## AI TECHNOLOGY FOR CLINICAL APPLICATION
AI technology has promoted clinical applications by improving patient outcomes, streamlining processes, and reducing costs. In clinical practice, AI has achieved striking success in analyzing patient data like brain-tumor segmentation,[53] assisting in clinical decision-making like epidemiological prediction,[54] and performing complex tasks such as surgery and rehabilitation, which indicates the potential to revolutionize healthcare service. In dentistry, the convolutional neural network has shown performance gain in detecting and classifying maxillofacial fractures from CT.[55] However, subtle details of maxillofacial fractures may not be accurately detected sometimes due to the unfavorable resolution of CT scans. Still, more advanced CT scanners can achieve higher-resolution images in future studies. Medical researchers also attempt to explore detection methods and investigate the feasibility of an automated decision-making tool for dental age estimation using deep-learning and topological approaches by analyzing the third molar maturity index (I3M) from 456 mandibular radiographs.[56] Another recent research proposed a more comprehensive AI system that can precisely identify individual teeth and alveolar bones from dental cone-beam CT (CBCT) images, which enables accurate and precise dental healthcare.[57]

The success of language modeling also promotes lots of research progress in representation learning for efficient medical services. For example, BioBert[58] is developed upon BERT to achieve a large but efficient text mining model for biomedical document analysis. ClinicalBert[59,60] carries out embedding training with a large volume of clinical documents to facilitate intelligent clinical diagnosis processes. SciBert[27] also built a large language model for representation learning with multiple documents across various scientific research domains.

One of the milestone contributions to biomedical research is AlphaFold,[61] developed by DeepMind. Its success in accurate 3D protein structure prediction demonstrates the power of large-scale training to tackle significant challenges in quantitative biomedical modeling. Since then, a series of innovative large-scale frameworks have been proposed to enhance AI-powered modeling, such as OpenFold[62] by OpenAI and BioNeMo Megatron[63] by NVIDIA, etc. In addition, inspired by LLM pretraining schemes, NVIDIA developed ProT-VAE[64] to advance functional protein design, which indicates the potential of large-scale biomolecule-language pretraining with an LLM. As technology continues to evolve, we can expect to see even more innovative applications of AI in clinical settings, ultimately leading to more effective healthcare services tailored to the needs of patients.

## ON EXPLORING THE CAPABILITY OF LLMS IN DENTISTRY
Automated dental diagnosis with an LLM
*Record analysis with text mining.* Contemporary medical practice widely adopts electronic health records (EHRs) for patient information documentation. Although it facilitates record generation and management, efficient analysis is still challenging since massive amounts of records are mixed with structured and unstructured data. This challenge leads to substantial amounts of underutilized data and obstructs patient care and research improvement. Text mining is able to tackle this challenge by drawing valuable conclusions and information from textual material in a mixed structure. Some straightforward modeling frameworks have been developed to capture patterns, correlations, and trends within textual data.[65–68] However, the performance of these models is inadequate to process massive amounts of documents efficiently and accurately.

An LLM helps to find a workaround for this limitation through training on extensive documents. Given strong competence in semantic understanding, an LLM can manage documents independent from structural formats. As shown in Fig. 1, text mining can also retrieve pertinent facts from unstructured data, such as free-text notes from healthcare professionals. From this unstructured data, an LLM like ChatGPT can be used to swiftly extract pertinent information, like a patient's unique illnesses or adverse effects.

*Treatment planning with natural language reasoning.* As aforementioned, medical service experiences the influx of a large volume of digital information. In addition to straightforward document analysis, these data can assist healthcare providers in customizing treatment plans.[69] Although this data analytics is stimulating, the work is taxing since more labor is involved in document analysis. LLMs can easily automate document comprehension and make treatment plan analysis feasible, which reaps the benefit of large-scale pretraining. Furthermore, billions of documents help LLMs cultivate the capability of natural language reasoning (NLR) to perceive contexts. This capability of NLR can assist dental practitioners more efficiently in establishing treatment plans tailored to patients' backgrounds.[70] For instance, NLR algorithms can examine adverse drug reaction (ADR) patterns linked to various dental procedures and drugs.[71] Sometimes, drug administration can lead to gum bleeding and severer diseases like bisphosphonate-related osteonecrosis. Dentists can modify their treatment plans to lower the likelihood of side effects by understanding the most typical ADRs linked to particular medications. We maintain that an LLM can facilitate this process and provide a case of this application in Fig. 2. It has been found that NLR may be used to identify comorbidities by analyzing patient records for common risk factors and symptoms, identification of ADRs,[71] drug safety surveillance,[72] and patient education.[73]

*Medical documentation with natural language generation.* In dental clinical practice, a synthetic yet faithfully representative EHR is essential to efficient medical information conveyance between healthcare providers and other medical professionals. Traditionally, this document preparation process is completed manually. Given keywords, practitioners organize the context following medical record documentation standards. It can be quickly assisted with an LLM. Natural Language Generation (NLG) is one of the important tasks of LLM. Generally, NLG accomplishes text generation given the understanding of natural language input, like structured texts or separate keywords. Since a well-trained LLM is highly skilled in this task, this merit can be leveraged to automate a series of common documentation tasks, such as generating reports on medical history, dental procedures, and treatment plans. For example, Fig. 3 shows an example of medical report narrative generation with only several keywords by ChatGPT. The generated texts are well formatted in a professional structure and encompass all important information in eloquent expression. The validity of the generated content has been validated.[74,75]

Compared to other application fields, the influence of AI in dentistry has unquestionably been slower and more constrained. This is mostly because patient privacy concerns have prevented patient data from being made widely available to the broader AI research community. Meanwhile, training data plays a crucial role in the advancement of AI methods. To address this dilemma and promote methodological advances in dentistry, using high-quality synthetic quasi-EHRs data is a practical approach since it facilitates data sharing between healthcare providers and external investigators. We employ ChatGPT 3.5, configured with specific parameters, to generate synthetic data, shown in Fig. 4. The parameters include max tokens, frequency penalty, and presence penalty, which were set to enhance diversity in the generated text. The frequency penalty reduces the likelihood of selecting words based on their frequency of occurrence, while the presence penalty imposes a fixed cost on each word in the text. These penalties encourage the model to generate text with higher perplexity rather than relying solely on the most probable word choices. Additionally, temperature scaling is used to adjust the distribution of probabilities for the next tokens, and a top-p value of 1 ensures consideration of all available tokens. Post-processing is applied to refine the generated data to eliminate any artifacts introduced during the generation process. These post-processing rules are determined through manual examination. These data can be quickly generated and harvested with the assistance of LLMs. Synthetic EHRs can be more realistic by introducing variability in the generated data. LLMs can be guided to generate different patient profiles, medical histories, treatment plans, and outcomes. This helps mimic the diversity and complexity seen in real EHRs. Thus, it implies that an LLM has competence in efficiently preparing medical information and protecting the privacy of patients.

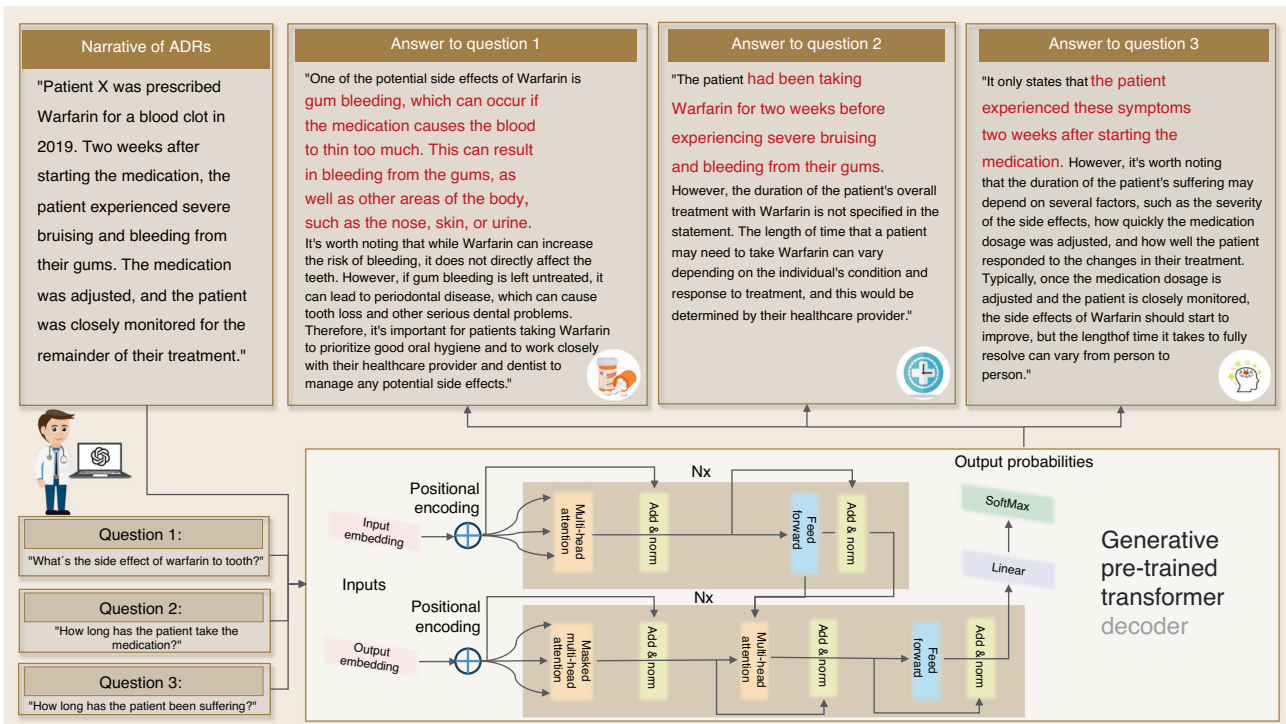

**Fig. 1** Examples of a text mining application for extracting a patient's historical record with an LLM like ChatGPT. Keywords are highlighted in red

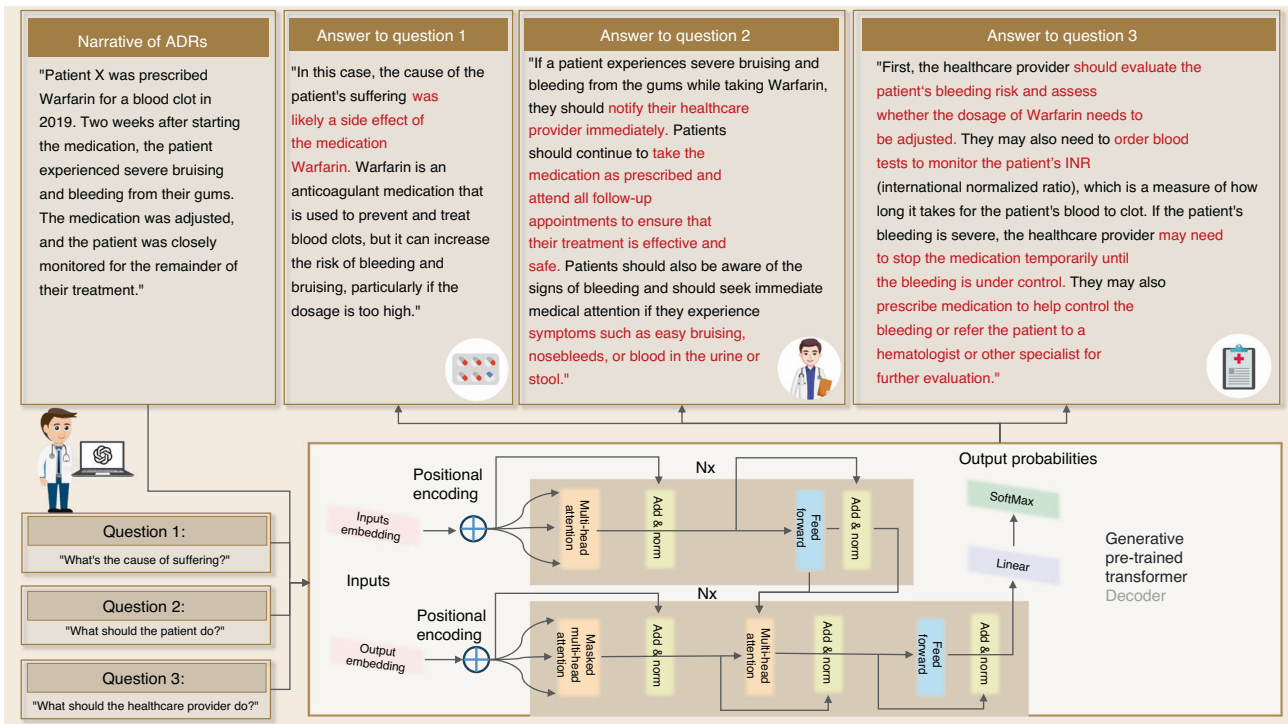

**Fig. 2** Examples of natural language reasoning (NLR) application of medication suggestions derived from a patient's record. Keywords are highlighted in red

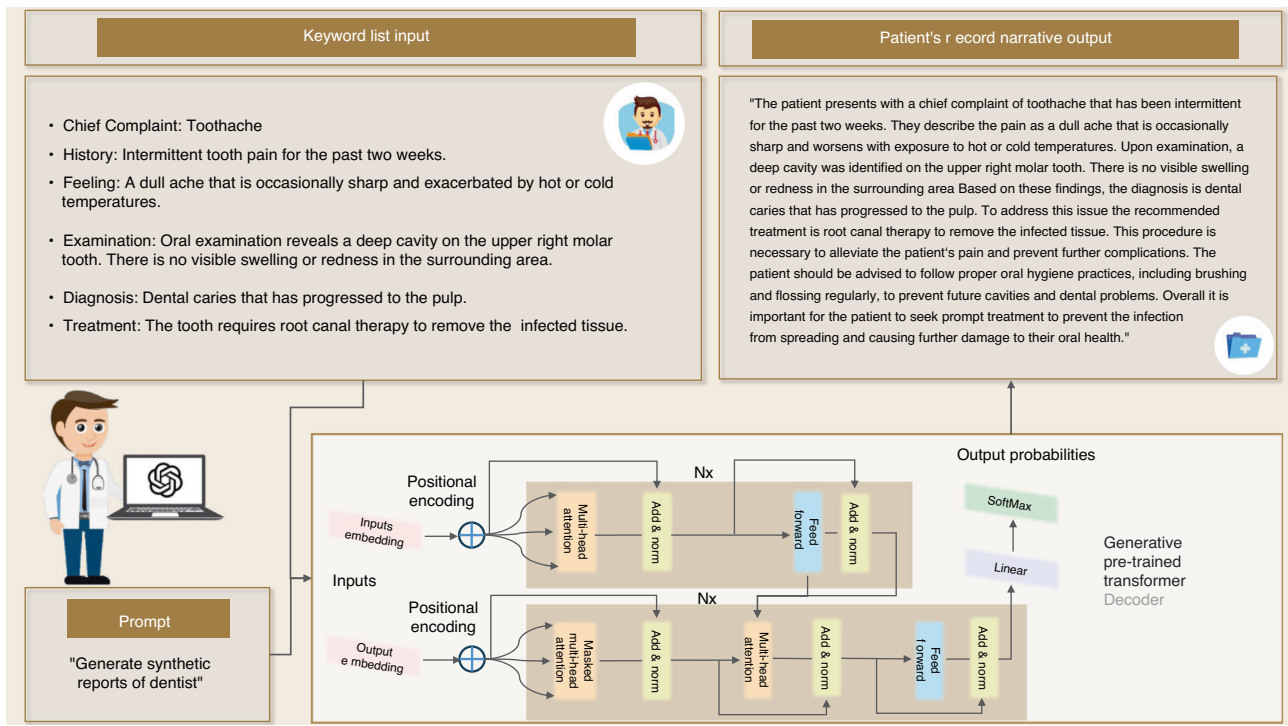

**Fig. 3** Example of a narrative output of the Patient's record generated from keywords with NLG

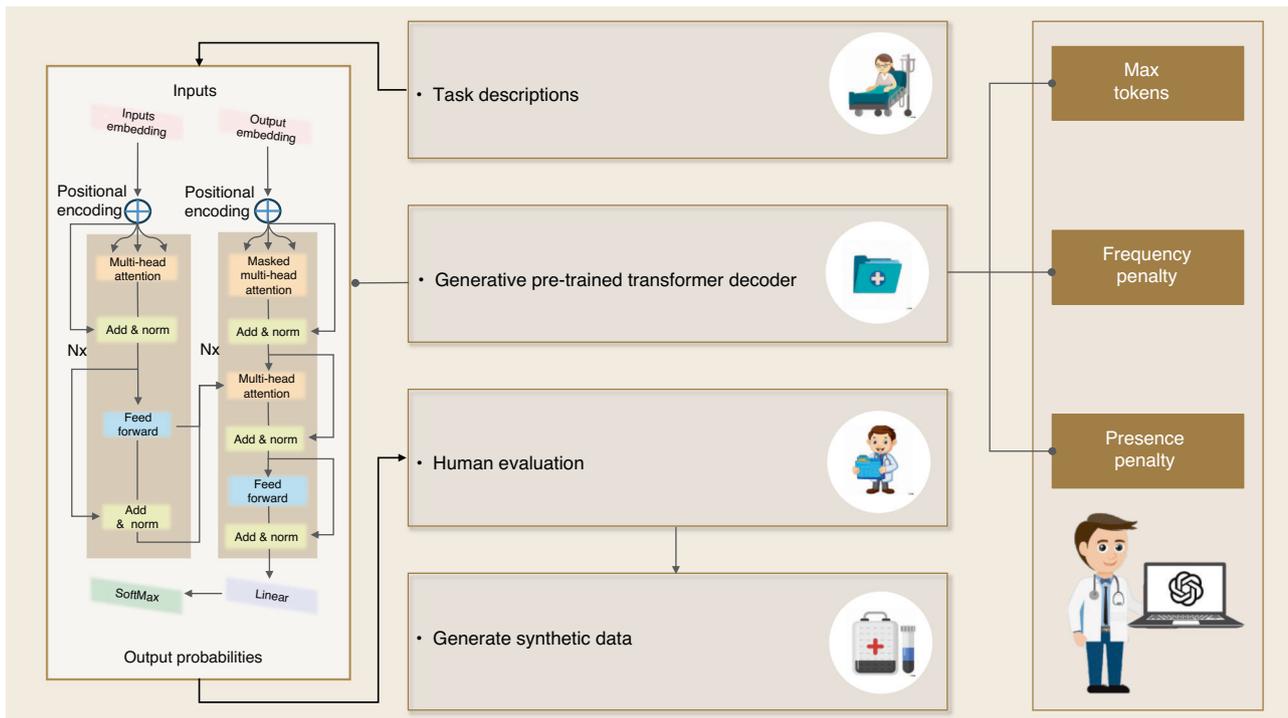

**Fig. 4**  Framework of generating synthetic quasi-EHRs data by LLMs

### Cross-modal dental diagnosis with LLMs

#### Vision-language deployment

Visual grounding:   Traditionally caries-related diagnosis is administered by dentists through visual and tactile examination. Before any treatment plan, a simple but comprehensive examination of oral health conditions is imperative. It sometimes takes experts' effort and time to diagnose tooth conditions, possibly through X-ray images and CBCT, and reach a reliable conclusion. Some existing research has explored the potential of AI-assistant models in assisting diagnose for caries,[3] periodontitis,[5] medication-related osteonecrosis,[76] maxillofacial bone fracture,[55] oral squamous cell carcinoma,[77] and temporomandibular disorders.[78] These diseases can be diagnosed based on medical imaging. Also, AI-assistant models for imaging analyses show the potential in assisting dental treatment, including orthodontics,[79] restorative dentistry,[80] oral implantology,[6] and oral and maxillofacial surgery.[7]

However, limited data representation hinders accurate diagnosis and treatment planning when the disease is intricate. The majority of study has been confined to image-only approaches, which restrain the effective conveyance of information and explore the untapped potential of AI models in dentistry. LLMs open the possibility of data-fused diagnosis by leveraging cross-modal perception. An LLM is highly skilled in aligning textual and visual representations for image-text analysis, which can facilitate the diagnosis of tooth problems by X-ray image interpretation.

Specifically, the inference by an LLM can be blended with specific visualization techniques to identify caries regions. For example, the practitioners can provide some keywords to query the model of ALBEF (A Lite BERT for Adaptive Embedding Factorization), which is specifically designed for image-to-text tasks and is integrated with Grad-CAM (Gradient-weighted Class Activation Mapping) to visualize the critical region for decision-making from the ALBEF model. The warmer color indicates the plausible regions corresponding to the described words. As shown in Fig. 5, root canal therapy is plausibly required in the region with a warmer color. Another tangible benefit of an LLM is training cost reduction. Without fine-tuning a large set of image data, an LLM can provide plausible affected teeth and likely locations of dental problems.

Visual question answering:   In addition to visual examination of medical imaging data, diagnosis documentation is more critical to patient-centered care. From the interpretation of X-ray images, healthcare professionals will write down the observation, analysis, and medication suggestions to patients. These documents are also essential to healthcare big data analytics, while the document summarization on the medical transcripts takes much time. An LLM is able to reduce the processing labor significantly through specific tasks, like visual question answering (VQA). Commonly, a VQA model can convert the encoded image representations to word embedding for dental diagnosis questions. With the diagnosis questions, the answers are generated to facilitate diagnostic report generation. This VQA-assisted diagnosis can be performed to assess potential dental health issues.[81] As shown in Fig. 6, the X-ray image of a patient's teeth is fed into an image encoder, like BLIP-2, generating a natural language representation, i.e., embedding based upon image understanding. Meanwhile, different questions are fed into LLM to generate another set of question embeddings. Both question and image embeddings are mathematically combined to generate the interpreted answers to the questions about images.

However, sometimes the raw image contains too much noise, or the resolution is not acceptable by the BLIP model; therefore, it is difficult to extract the desired information through the VQA model. To address this, training a semantic segmentation model that divides the areas of the image with different properties into different classes is a potential solution because it allows an LLM to learn each element separately.[82] For example, as Fig. 7 shows, the soft tissue envelope and nasal septum/concha are classified into orange and blue segments, respectively, and it is expected to improve the model performance and enhance image understanding to extract the morphology information of nasal cartilages, as the cartilages are small and embedded by the soft tissue.[83] Figure 7 also demonstrates the potential of training machine learning models to help reconstruct the nasal cartilage based on MRI for patients with orofacial clefts, who can suffer significant nasal deformity.[84–87] Due to the limitations of current imaging software, the differences between the cartilage and soft tissue cannot be easily defined, but with training machine learning models, the purpose should be achieved in the future.

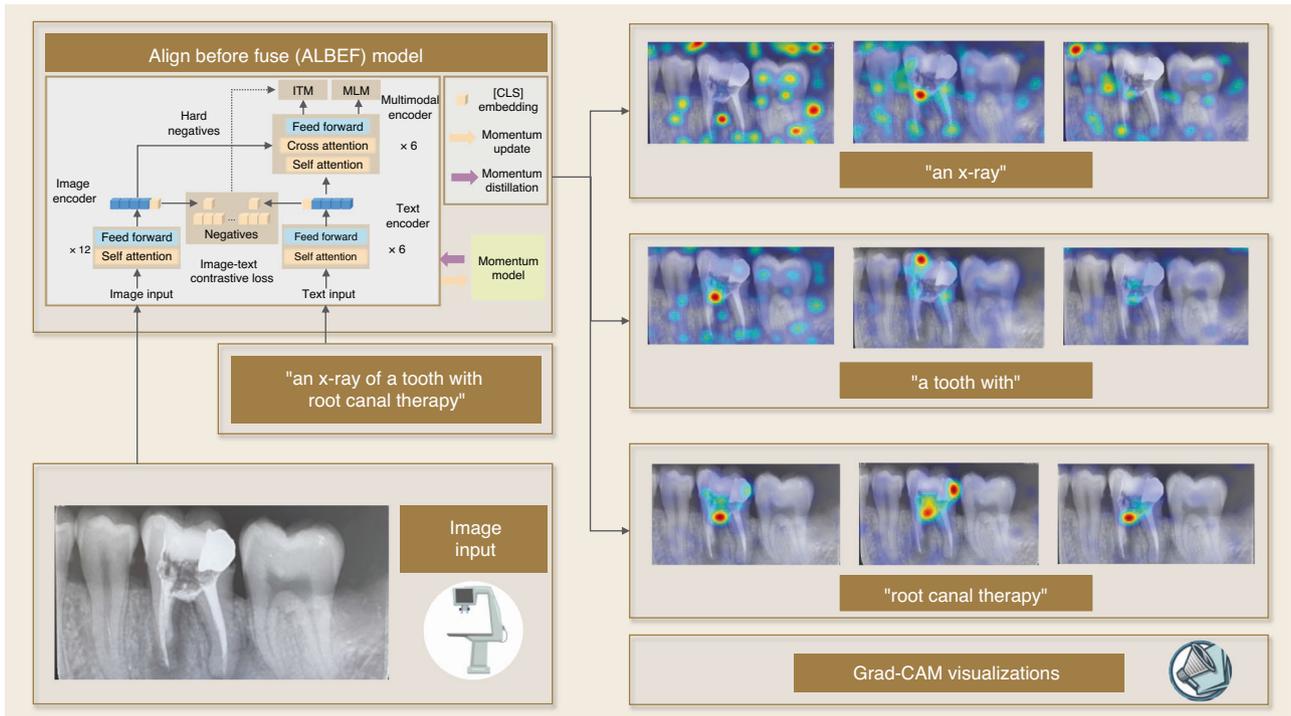

**Fig. 5** Schematic of dental condition diagnosis with a vision-language model of ALBEF

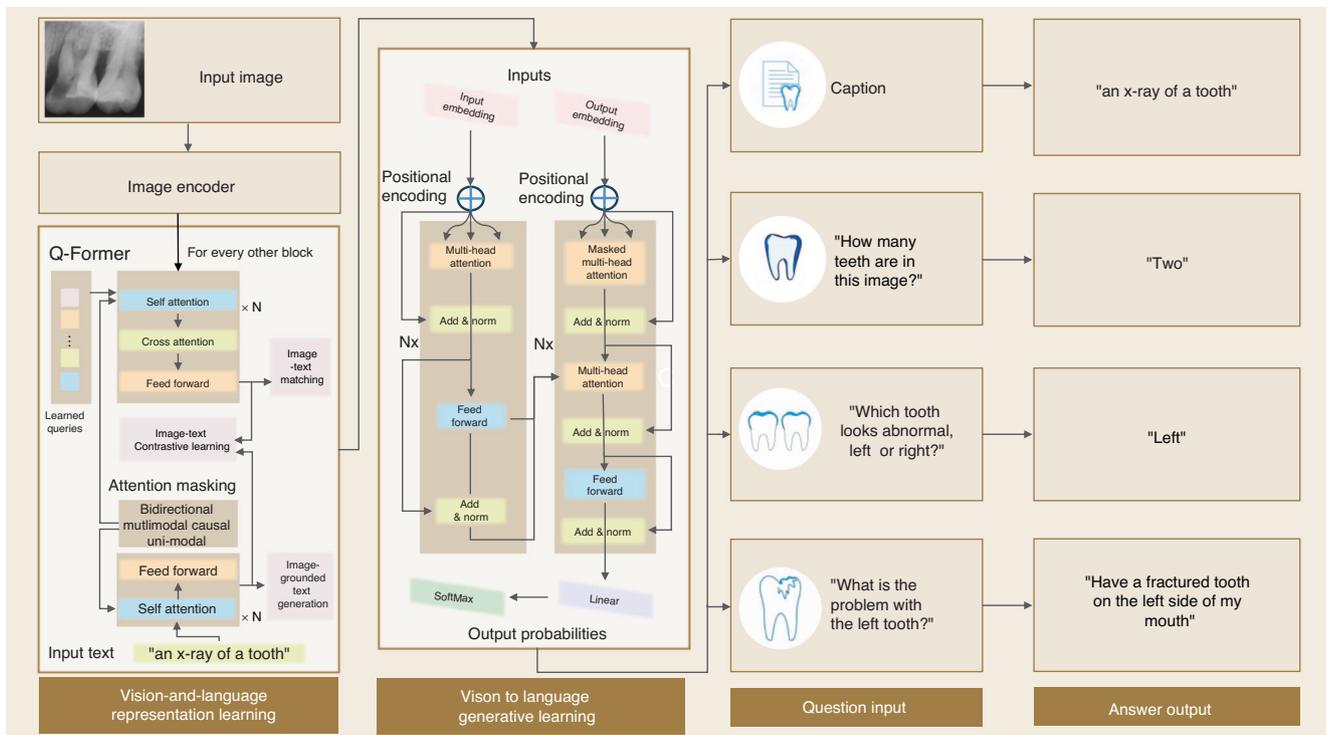

**Fig. 6** A VQA example framework with the assistance of BLIP-2

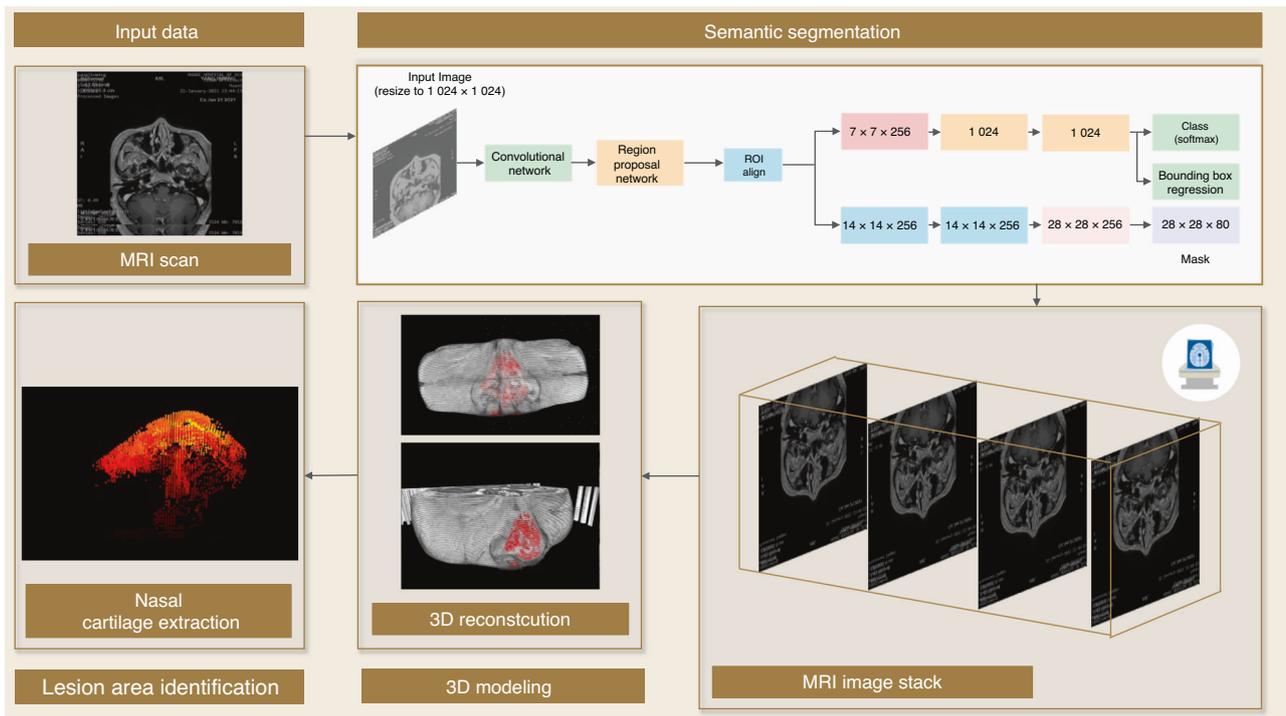

**Fig. 7**  2D semantic segmentation with 3D reconstruction for lesion identification

Visual data generation: DALL-E 2, empowered by integrating GPT-3.5 as encoder and iffusion model as decoder, can be utilized in the medical field to generate synthetic medical images. Once the LLM has been trained on real EHR data, it can be used to generate synthetic quasi-EHRs. By providing prompts or specific instructions to the model, such as patient characteristics or medical conditions, the LLM can generate realistic synthetic records that resemble real-world EHRs. For example, by describing a patient's CBCT scan in text, including details of any abnormalities, like an odontogenic cyst or alveolar cleft, which can observe obvious lesions on the alveolar bone structure, and feeding it into DALL-E 2, synthetic medical images that match the description can be efficiently produced in large quantities to improve the performance of deep-learning models by use as the training dataset. Figure 8 shows that synthetic medical images can be generated with varying levels of noise, contrast, or resolution to create images with specific properties or characteristics that are relevant to the medical condition being studied. Moreover, patient privacy can be protected since synthetic medical images are generated from textual descriptions rather than real medical data. This technique is valuable for medical research based on any medical 3D imaging techniques, including CBCT, CT, MRI, etc., and improving patient care by training machine learning models while maintaining patient privacy.

Alternatively, LLMs can also be employed to generate medical illustrations or diagrams based on textual descriptions. For example, a description of a surgical procedure can be fed into illustration software to create an accurate and detailed illustration of the procedure.

*Audio-language deployment*. Besides imaging and dialogs, a patient's voice is also critical to medical diagnosis. A person's voice can potentially reveal important clues about their speech function, as certain vocal characteristics may be indicative of the function of teeth, tongue, pharyngeal structure, and muscles. Analyzing these vocal attributes can assist healthcare professionals in identifying potential health concerns. One of the common medical diagnosis applications is waveform-spectrogram analysis on patients' audio recordings, which are collected by requesting the patients to read certain words or paragraphs. The waveform is a curve-based representation of an audio signal, the shape of which enables acoustic analysis. The spectrogram is an alternative representation of sounds in the frequency domain, which facilitates signal processing and analysis.

Velopharyngeal insufficiency related to cleft palate, oronasal fistula, and so forth, that affects the contaction between the soft palate and posterior pharyngeal wall, or changes the needed separation between the oral cavity and nasal cavity,[87–89] exhibits some typical marks on voice waveforms and spectrograms. In velopharyngeal insufficiency, for example, nasal emission can lead to distinct variation in speech.[90] A person with velopharyngeal insufficiency may exhibit a waveform that shows the reduced intensity of the sound waves during certain frequencies or periods, leading to altered speech patterns because of the abnormal airflow in oral and nasal cavities.[91–94] Figure 9 shows an example of a comparison of the waveforms and spectrograms between normal people and patients with velopharyngeal insufficiency. It can be observed that normal people have a more intense waveform and continuous spectrogram, while the patients' samples are more dispersive and broken.

Furthermore, the waveforms and spectrograms of different patients can be fed into the pretrained LLMs such as GPT-4 for potential disease and severity diagnosis. As shown in Fig. 10, a pair of graphs are inputted into GPT-4 while asking for disease deducing, and the model provides several answers for reference. Although the final answers show little about the velopharyngeal insufficiency, the output mentions muscle dysfunction. Also, a more precise output can be achieved by further fine-tuning with more labeled patients' audio data. In addition, NLG can also be used in conjunction with speech recognition software to convert voice commands into written text, such as when dictating clinical notes or treatment plans.

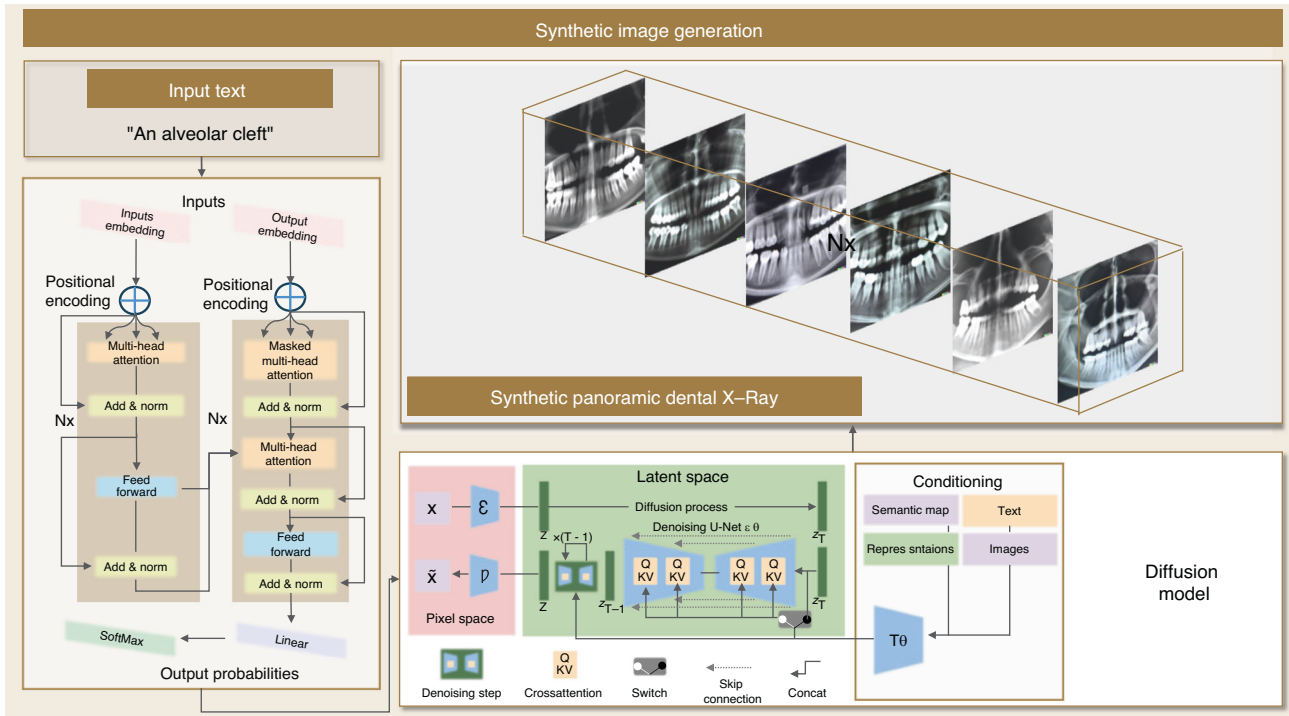

**Fig. 8** Example of visual data generation

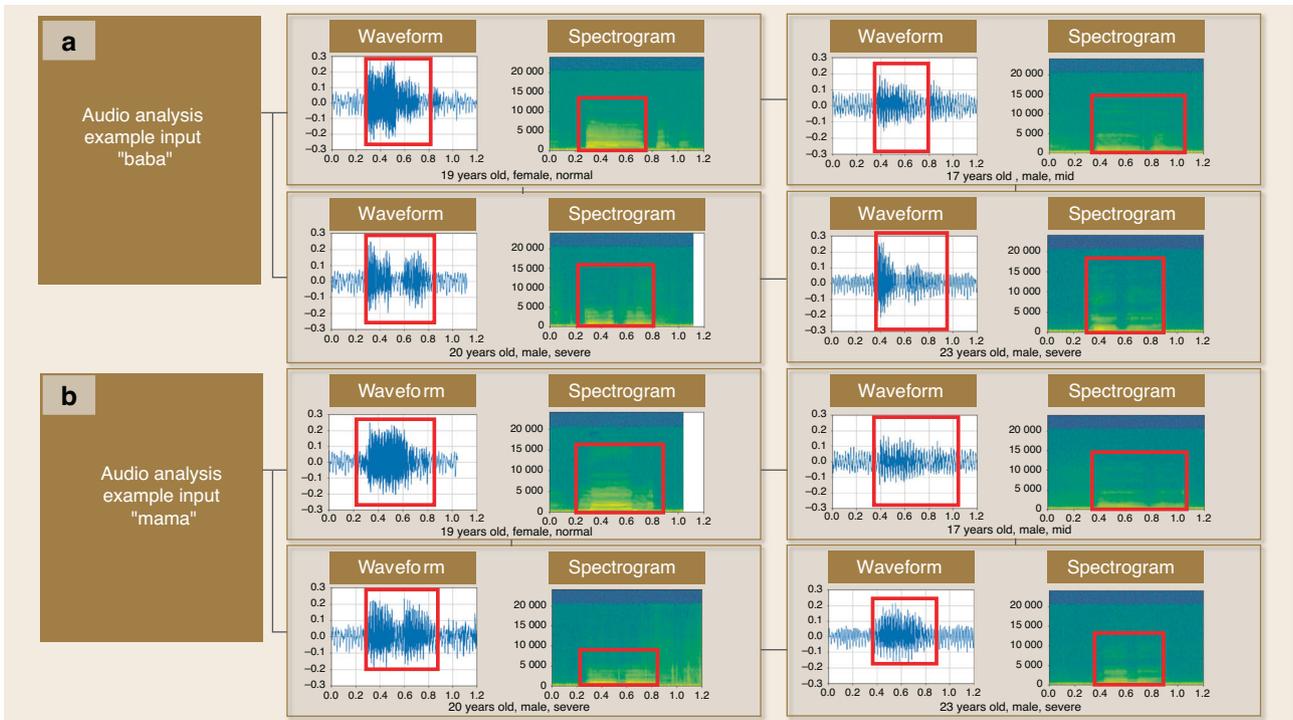

**Fig. 9** Example of the audio waveform and spectrogram analysis in speaking **a** "baba" and **b** "mama" with TorchAudio of normal people and patients with velopharyngeal insufficiency

## OTHER POTENTIAL CROSS-MODAL DEPLOYMENTS

### Biopsy

A biopsy is a medical procedure in which a small sample of tissue is removed from a person's body to be examined under a microscope. This is typically done to diagnose or investigate various conditions, such as cancer, infections, or inflammation. Visualization techniques can be applied to biopsy to understand the tissue structure and cell morphology better and identify any abnormalities. Common visualization approaches include light microscopy, immunohistochemistry (IHC), immunofluorescence (IF), confocal microscopy, etc. With the embedding generated by a vision transformer, the image input can be projected into the language space and used for understanding the characteristics of disease identification. For instance, prostate cancer in biopsies[95] and pre-implantation kidney biopsy pathology practice[96] has been regarded as potential application fields, which also shows the potential for biopsy and histological analyses in dentistry, and oral and maxillofacial surgery.[97,98]

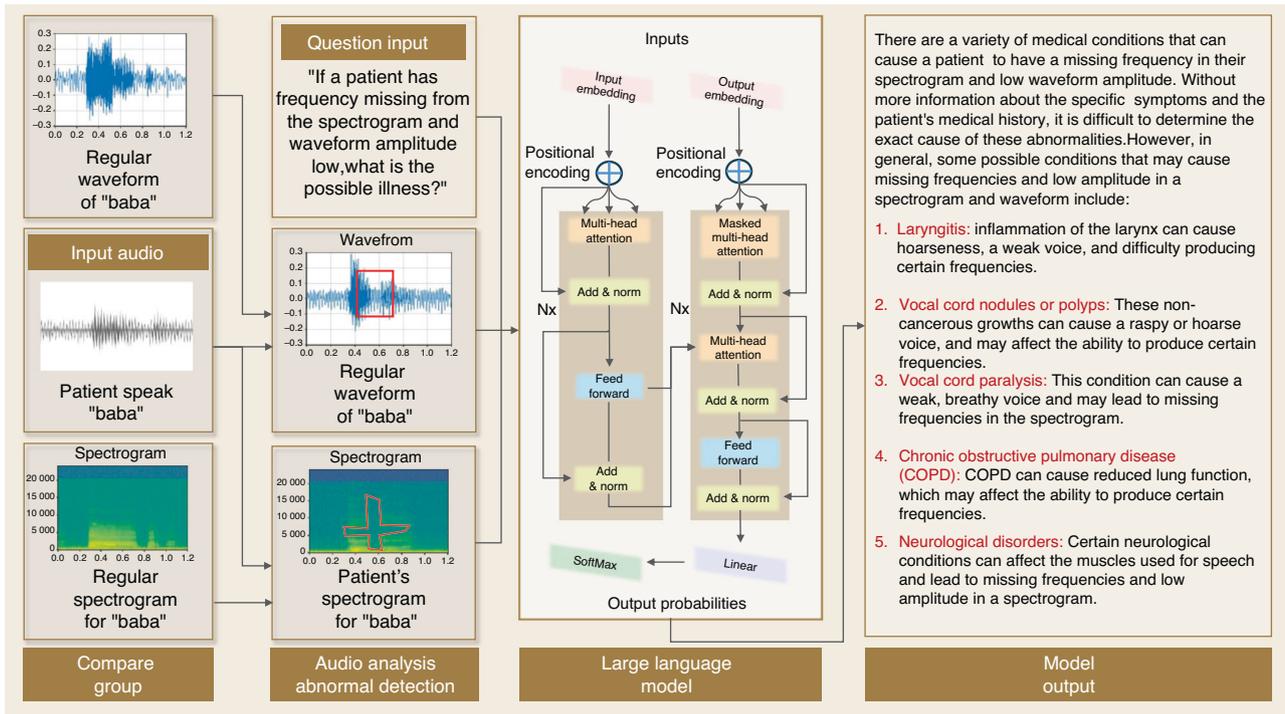

**Fig. 10** Schematic of audio-language assisted diagnosis based upon audio waveform and spectrogram analysis with TorchAudio

Blood test
LLMs can help users understand the results of their blood tests by providing information on the normal ranges for different biomarkers and explaining the potential implications of high/low values. The changes in the test indicators also provide rich information about the human body condition, as well as track the recovery or disease deterioration process. These changed conditions can affect the patient's treatment planning and treatment for dental problems.[99] For instance, anemia may present with low hemoglobin, hematocrit, and red blood cell count, while liver disease may present with elevated liver enzymes (ALT, AST, and ALP). These abnormal parameters may postpone the treatments like oral and maxillofacial surgeries as most of these surgeries are elective. The internal relationships and connections between these indicators can be well captured by LLMs, and thus potential diseases can be linked with the inputted information.

Gene detection
Gene detection is the process of identifying and analyzing specific genes or genetic sequences. Classic or more recent approaches can help obtain the gene sequence, including RNA sequencing, DNA sequencing, single-cell sequencing, etc. The genes or genetic sequence can be projected into language embeddings with the corresponding encoder and then input to LLMs. As the sequence contains the underlying logic of the gene's property, the LLMs can help to understand these logics by learning from large gene samples after training.[100,101] Potential applications may include understanding gene function, identifying genetic variations or mutations, and studying the relationships between genes and various biological processes or diseases, which can further influence the development of dental problems and treatments related to genetic disorders.

## AI SYSTEM FOR DENTISTRY APPLICATION WITH A FULLY AUTOMATIC MULTI-MODAL LLM
To demonstrate the effectiveness and potential of LLMs' application in dentistry, we present a framework of a fully automatic diagnosis system based on Multi-Modal LLMs. The system consists of three input modules from different models: vision input, audio input, and language input.

The image input could be dental X-ray, cone-beam computed tomography, and other medical imaging. For semantic classification, we focus on optimizing the capture of the critical elements. By applying vision-language models, the condition of the tooth is evaluated, potential anomaly or disease is detected, and specific diagnosis and corresponding suggestions can be given.

In this case, audio sources have two usages: voice anomaly detection and patients' narrative understanding. For the first usage, the system receives the voice input from patients, plots waveform and spectrogram, and then performs amplitude and frequency analysis. For the second usage, the patients' narratives are collected and converted into texts using speech recognition techniques. Afterward, the key elements, like the symptoms that patients stated, can be extracted and summarized to form reports or bullet points for doctors' reference.

Targeting automatic diagnosis for dentistry, the AI system can be embedded into the dental clinics' internal communication systems. Thus, a fully developed automatic application can encompass patients' information from multiple sources and provide a professional medical diagnosis, as shown in Fig. 11.

## A SPECIFIC CASE FOR THE MULTI-MODAL LLM AI SYSTEM FOR DENTISTRY CLINICAL APPLICATION
To demonstrate the application of the multi-modal LLM AI system in dentistry, we use a sample with dental caries to explain how it works by LLM, including vision-language modeling and treatment planning with natural language reasoning. As shown in Fig. 12, an X-ray of the tooth is inputted into the system, and the abnormal morphology, like decay on the tooth, can be located on the X-ray by vision-language modeling, and then the first question can be answered that dental caries can be found on the tooth. Then the next question becomes what the treatment planning for this problem is, and using the LLM again to output seven steps, including communicating with the patient, developing a treatment plan, discussing treatment options, conducting necessary dental procedures, providing oral hygiene instruction, scheduling follow-up appointments, and emphasizing preventive measures. However, from the X-ray, we can also observe potential bone loss near the distal root, which is not detected by the system according to this pilot study. Thus, further study should be done to improve the system.

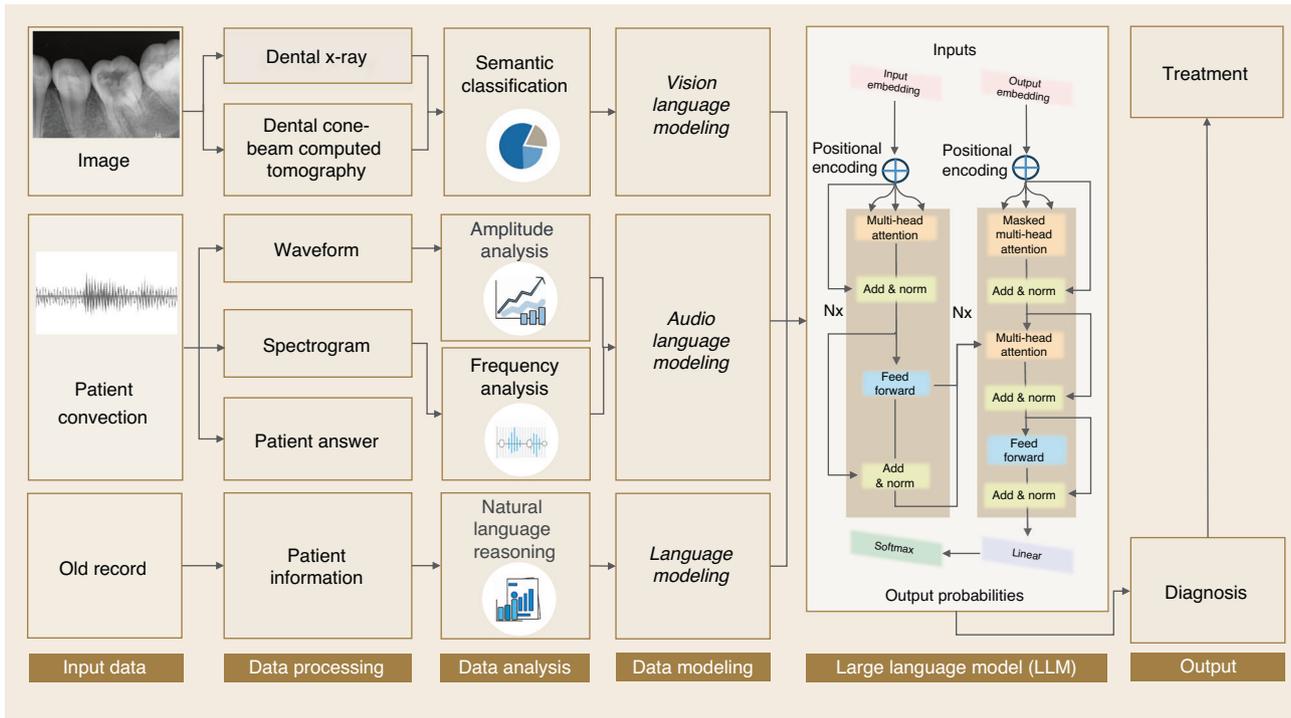

**Fig. 11** Concept of automatic multi-modal LLM AI system for dentistry application

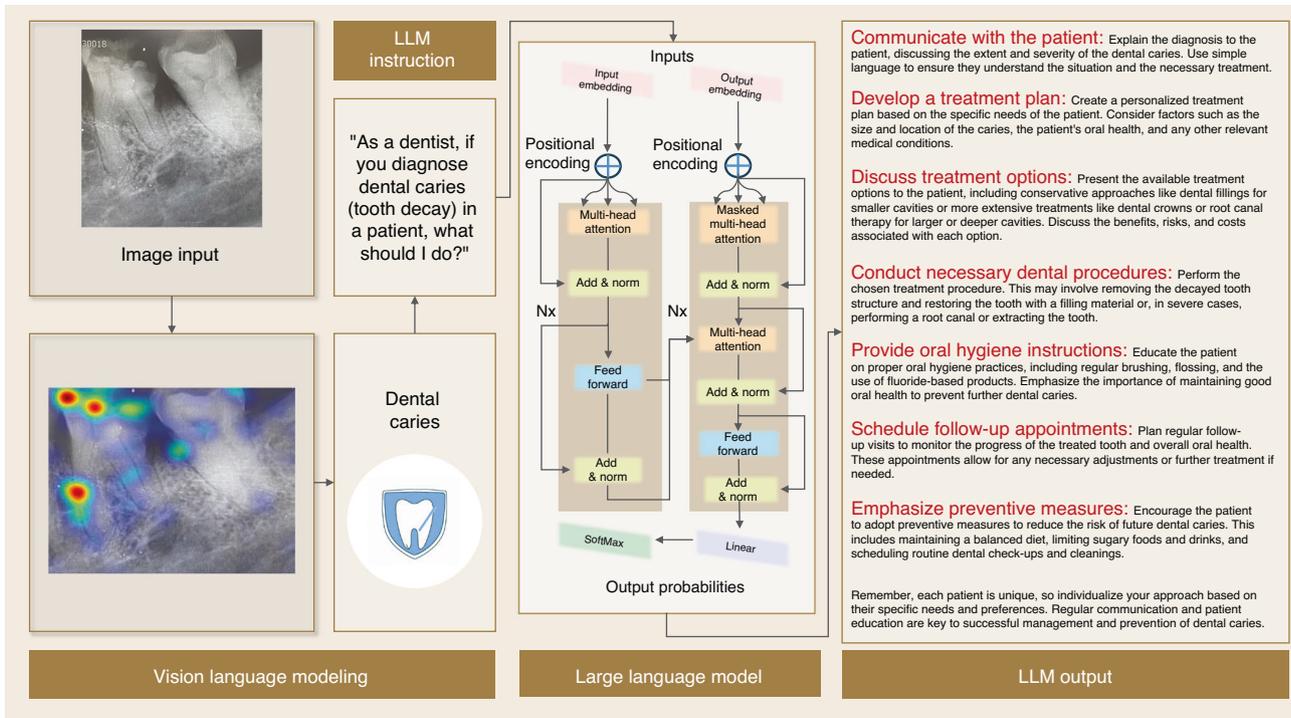

**Fig. 12** Application of the multi-modal LLM AI system in dental caries

## ISSUES AND LIMITATIONS

While there is much excitement around the potential applications of LLMs in the field of dentistry, some issues and limitations must be addressed before these models can be widely adopted.

### Data quality

Despite rigorous efforts to sanitize and filter the vast amount of training data, it is challenging to eliminate all harmful or inappropriate content, which may inadvertently propagate through the responses generated by LLMs.[102] Inherently, these LLMs operate as sophisticated pattern-matching machines without a genuine understanding of the data they are trained on, which occasionally leads to nonsensical or inappropriate responses.[103] Compounding these issues, they lack the capacity to validate the information they generate, remaining incapable of accessing real-time data or verifying the current status of events post-training. Moreover, the knowledge base of an LLM is static, established at the time of training, thus unable to update its knowledge or assimilate new developments in the evolving data landscape. One possible solution could be to develop "Human-in-the-loop system". Pairing LLM systems with human supervisors can safeguard important decisions, helping to catch and correct mistakes that the LLM might make[104].

### Model bias

Biased clinicopathologic analysis results are the first noteworthy issue. Because LLMs are data-driven-only models which learn the features and patterns in the training data, the correctness of the LLM is highly dependent on the quality and adequacy of the data.[105] Although the LLMs are evolving iteratively, even in the era of GPT-4, we can't fully trust the AI-generated clinicopathologic analysis results, and human-in-the loop validating work is still necessary. In the future, neural-symbolic models, which can combine two approaches (neural networks and symbolic reasoning) by using neural networks to learn the statistical patterns in large datasets and then using symbolic reasoning to perform logical operations on the learned representations, can be a potential research direction.[106]

### Data privacy

The patient data breach is another big issue, especially in today's privacy-sensitive world. Fine-tuning the LLMs in the dentistry domain is expected, and a huge amount of patient data is necessary. A data breach is likely to happen during this process if the healthcare providers and developers don't take appropriate measures to safeguard patient data. It's crucial to implement strict data handling protocols and use secure communication channels for transmitting and storing patient data. In addition to these security measures, it is equally essential to inform patients about using their data in advance and obtain their consent.[107] Another possible point of data leakage is dental diagnosis. Inputting patient data is necessary for these diagnosing applications, where there is a risk of violating patient privacy and confidentiality. One possible solution to address this concern is to use offline LLM such as META LLaMA, where the LLM is run locally on the device or edge server rather than on a centralized server API Call.[108]

### Computational cost

Computational resources can also limit the application of LLMs in dentistry. It is reasonable to expect that the LLMs in the dentistry domain will be operated locally due to data sensitivity. Firstly, fine-tuning the LLMs in the dentistry domain using local computational resources can be challenging. Then, running a full LLM to support the application in the dentistry domain is a waste of computational resources and unnecessary. A sparse expert model, a type of LLM that incorporates a set of specialized expert models, can be a future solution. It can reduce the computational resources required to train and run LLMs while handling specific tasks or domains more efficiently than the main LLM.

## CONCLUSIONS

The utilization of language models like ChatGPT holds significant potential for advancing clinical applications and research in dentistry. By employing these models in a rational manner, a paradigm shift can be achieved in dental diagnosis and treatment planning. Further exploration based on diverse medical examination data will facilitate the realization of precision medicine and personalized healthcare in dentistry. A crucial future endeavor of practical deployment involves fine-tuning language models with dentistry domain-specific knowledge. This entails training the models with dentistry teaching materials, patient records, and other relevant domain information, resulting in enhanced accuracy by capturing pertinent patterns, terminology, and context. Consequently, the models acquire a profound comprehension of dentistry concepts, enabling them to generate contextually relevant and insightful responses. Customizing the outputs in alignment with domain requirements and preferences enhances efficiency, saving valuable time and resources. These benefits substantially contribute to improved performance and usability, rendering fine-tuned language models invaluable tools for research paper composition. Concurrently, the adoption of LLMs will further reduce medical costs and enhance medical efficiency.

## AUTHOR CONTRIBUTIONS

H.H. and O.Z. contributed to the conception and design of the work. H.H., O.Z., D.W., J.Y., Z.W., S.D., H.Y., C.X., and R.Y. performed data acquisition, analysis, and interpretation. O.Z. and D.W. performed data acquisition. H.H., O.Z., and D.W. performed analysis and interpretation. H.H., O.Z., D.W., J.Y., Z.W., S.D., H.Y., C.X., R.Y., Q.Z., and B.S. drafted and critically revised the paper. All authors gave final approval and agreed to be accountable for all aspects of the work.

## FUNDING

This work was supported by the Research and Development Program, West China Hospital of Stomatology, Sichuan University (RD-02-202107), Sichuan Province Science and Technology Support Program (2022NSFSC0743), and Sichuan Post-doctoral Science Foundation (TB2022005) grant to H. Huang.

## ADDITIONAL INFORMATION

**Competing interests:** The authors declare no competing interests.